# Specific polysemy
# of the brief sapiential units

Marie-Christine Bornes Varol
département Etudes Hébraïques
Inalco Paris. France

Jean-Daniel Gronoff
www.DualSemantics.Org Paris. France

Marie-Sol Ortola Querol
Université Nancy2 Nancy. France

**Abstract**: In this paper we explain how we deal with the problems related to the constitution of the Aliento database, the complexity of which has to do with the type of phrases we work with, the differences between languages, the type of information we want to see emerge. The correct tagging of the specific polysemy of brief sapiential units is an important step in the preparation of the text within the corpus which will be submitted to compute similarities and posterity of the units.

## Introduction

Proverbs form an essential part of the oral and written culture of the Mediterranean basin since the oldest antiquity. They represent a specific type of philosophical knowledge. Many of these sayings, maxims and proverbs are still used nowadays.

The aim of the project Aliento[1] is thus the study of proverbs, sentences, maxims... from the Iberian Peninsula (IXth - XVth C.), their circulation between three cultures and several languages, their sources and their posterity. The spatial area has been chosen for its central role in the circulation of knowledge during the Middle Ages. The database will establish how transmission occurs from the ancient sapiential sources to the modern and contemporary Mediterranean books of proverbs. This is why we must cross and intersect texts from distinctive cultural traditions, written in different languages, annotated collectively in a uniform way. The database assembles and compares digitalized sapiential texts in Latin, Arabic, Hebrew, Greek and Romance languages and it is build and accessed by the way of the collaborative website 'Aliento.eu'. Interrogation languages are Spanish, French and English.

The extension and variety of the texts we deal with require a critical pinpointing and a precise classification of the relevant brief statements. The themes of the discourses are sapiential from beginning to end. However, what we want to see emerge is the brief sapiential units only, that is to say, a subgroup within a textual corpus. Each unit is meaningful within a context or a succession of statements from which it can't be separated. These brief units evolve depending on particular temporal periods; they become proverbs, independent statements or not.

---

1   The project Aliento (linguistic & intercultural analysis of brief sapiential statements of the Iberian Peninsula (IXth – XVth C.) and their transmission Orient / Occident & Occident / Orient) is a project from MSH Lorraine, INALCO – Paris, Université Nancy2, Belunet & DualSemantics.Org.

Relevant operating parameters exist, such as the nature of the brief sapiential unit, the language and the culture which it belongs to, its dating, the geographical area which it comes from (from within the Peninsula or outside).

This information is brought by textual annotations (concerning manuscript, critical edition, and so on) relative to the language of the text, its origin, its author, its dating, all this data coming from the scholarly tradition.

The text being taken as a whole, we must build a subgroup of comparable units. An essential work is to set apart the statements to be tagged, for their relevance in terms of their structure and their significance.

We must also take into consideration the fact that a statement found in our texts becomes a "proverb" in one culture, and is not yet such a thing in another, but it belongs to a lengthy expression containing an advice or an edifying discourse.

It is important to indicate that we have decided to tag differently the different parts of the text, at this stage of our study. The narrative parts of the text left aside are indicated by the tag: <text.del>; the eventual exempla: <exemplum>; similarities: <simil>; edifying discourse: <de>. We don't rule out the possibility of coming back to them. We have categorized in a different way the sapiential statements, depending on their degree of proverbialisation: <pr?> for the units potentially proverbial; <pr>: good candidates to proverbialisation; <pr.est>: established proverbs.

To compute similarities of meaning between the multilingual brief sapiential units of the database and to compute their posterity, the knowledge representaion is based on embedded XML annotations in order to represent the specific "polysemy" of each brief sapiential unit. XML annotations allow for possible reexamination depending on the results of the research, and for tracing back the evolution, the freezing process, the puns on set expressions, the analysis of the sapiential units set in a cultural, diachronic, synchronic perspective, and so on.

**Building a computational knowledge container for each brief sapiential unit**

To show how we build the knowledge representation of the specific polysemy of brief sapiential units, we used the following list of five brief sapiential units extracted from five texts belonging to the tradition assembled in our corpus of reference:

1. (Arabic) : وقد كنت مرتفعاً فأصبحت متّضعاً.
2. (Spanish_1) : E ante érades mucho alto, e agora sodes mucho baxo.
3. (Spanish_2) : Quanto fue alta la su sobida tanto fue mas baxa la su cayda.

4. (Latin)　　　: Quanto altitudo Alexandri excellencior fuit, tanto gravior est casus.

5. (Hebrew): ‏. והיית נכבד והינך שפל‏

The units of the list are collected by specialists to correspond to the same kind of sapiential lesson. They belong to a group of texts closely related, as shown by scholars of the different fields of studies, which interest Aliento project directly.

This is the list of the English literal translation of the five brief sapiential units:

1. and you were on the top and you are down
2. And before you were very high, and now you are very low
3. the higher was his hight the lower was his fall
4. greater was Alexander's hight, more painful is his fall
5. And you were an eminent person and now you are down

We first link each unit to a meaningful set of information, such as translations of the literal and figurative meaning in French, English and Spanish (languages of interrogation of the database), junctions to nowadays used proverbs, which represent the principal form of questioning in current paremiological databases; the "romanisation" of the units written in languages with characters other than latin. Romanisation is a system of simplified transliteration, which avoids diacritic signs, for example. Difficulties inherent to Semitic languages graphical output have been jointly solved by Arabists and Hebraists, all members of Aliento. In order to compare units and to compute them as 'similar', we link each one to a number of information. All added information is represented as an XML entity linked to the appropriate unit.

This is for example the list of the XML entities linked to the unit "E ante érades mucho alto, e agora sodes mucho baxo" at the beginning of the process of comparison. We would like to emphasize at this point that the tags are specific to the Aliento project and that the encoding in such a way is made to create semantic clusters that will be analyzed by algorithmic calculation.

**\<pr.all>**
**\<pr>**e ante érades mucho alto, e agora sodes mucho baxo**\</pr>**
**\<pr.type.es>**cuanto más alto subas mayor será la caída**\</pr.type.es>**
**\<pr.type.fr>**…plus dure sera la chute**\</pr.type.fr>**
**\<sl.fr>**et avant tu étais très haut, et maintenant tu es très bas**\</sl.fr>**
**\<sl.en>**And before you were very high, and now you are very low**\</sl.en>**
**\<sf.fr>**Vivant vous étiez au sommet, mort vous êtes en dessous de tous**\</sf.fr>**
**\<sf.es>**Vivo estabas en la cumbre, muerto estas debajo de todos**\</sf.es>**
**\<sf.en>**Alive you were above all, dead you are below everyone**\</sf.en>**
**\<lec.fr>**La mort réduit le puissant à rien**\</lec.fr>**
**\<lec.es>**La muerte aniquila al potente**\</lec.es>**
**\<lec.en>** death annihilates the mighty**\</lec.en>**
**\<key.fr>** puissance déchéance mort chute **\</key.fr>**
**\<key.es>** potencia decadencia caída muerte **\</key.es>**
**\<key.en>** might decay fall death **\</key.en>**
**\<lem.es>**y antes ser muy alto ahora ser muy bajo**\</lem.es>**

```
<str.ling>Sadv1 SV Sadv2 / Sadv opp à 1 SV Sadv opp à 2</str.ling>
<str.form>parallélisme binaire : avant/après haut/bas </str.form>
<str.poet>8 / 8 ass a/o</str.poet>
</pr.all>
```

**<pr.all>**: the pair of tags (<pr.all> at the top of the list and </pr.all> at the end of the list) includes all the *XML* entities wich refer to the same brief sapiential unit. <pr.all> builds a kind of knowledge container where a lot of information linked to the brief sapiential unit can be freely included. The brief sapiential unit is the first XML entity of the container.

**<pr>**: the brief sapiential unit. This unit must be unique at the scale of the project. The unicity can be computed and certified later.

**<pr.type>** characteristic proverb or expression which is directly linked to the studied unit (all languages can appear but the main languages are the three languages of interrogation of the data base)

**<sl.fr>**: literal translation of the brief sapiential unit (<pr> entity), written in Spanish, into French.

**<sl.en>**: literal translation of the brief sapiential unit into English.

**<sf._>**" stands for figurative sense, and the suffix '._>' refers to the language of the entity ('.en>' '.fr>' '.sp>').

**<lec._>**" stands for sapiential lesson, and the suffix '._>' refers to the language of the entity ('.en>' '.fr>' '.sp>').

**<key._>**" stands for ideological keywords, and the suffix '._>' refers to the language of the entity ('.en>' '.fr>' '.sp>').

**<lem.es>**: the root "<lem" stands for lemmatization the language of the sapiential unit (Spanish).

Lemmatization is the traditional way for human specialists to compare multilingual entities. It can be used also for multilingual computed comparisons when it is an XML entity of the container.

Why do we use lemmatization? : Terms spelling and material forms vary even within the same language or in a same text, because spelling is not normalized neither fixed yet. Words can be as well characterized by an important phonetic and morphologic variation: "eres, sodes, sos" are the verb to be (ser) 2[nd] pers. sing. If we add to this the possibility of person, mode, time variation, lemmatization becomes very important and uncovers a number of homological ambiguities with terms that are in no way related.

Aliento's linguists solve graphic and morphological variation through lemmatization, which consists in the linking together a graphic form of the text (a word) and a "lemma", that is to say, an unmarked form of the word, a sort of "neutral basis", or a stem without any temporal, aspectual mark, and no distinction of voice, number, gender and syntactic agreement. This resolution leads to the constitution of an alphabetical dictionary of lemmas which will help find all the occurrences of a word in all the texts with all its morphological and orthographical variants.

For Aliento, the XML entity container represents a complex lexical encoding done in a well-thought and efficient way in order to simplify the computing of lexical categorization. It should, however, be mentioned that, since this categorization functions at the level of the word as an autonomous entity, it has a limited relevance at certain levels of the calculation.

**<str.ling>**: the root '<str.' stands for structure and the suffix '.ling' stands for linguistic. This entity uses a specific formal convention to describe the linguistic structure of the sapiential unit.

**<str.form>**: we indicate that we are dealing with a dialogue form (question/answer), an imperative sentence (exhortation), a parallel sentence structure, and so on.

Formal structures vary even if a certain number of them recur and are identifiable. This is the case of "si…" ("if…" or "whether…"); "como" with the meaning of "as" , "such as" or "like"; "quanto mas… tanto mas" ("the —er [the more]…, the—er…");

**<str.poet>**: the root '<str.' stands for structure and the suffix '.poet' stands for poetic. This entity uses a specific formal convention to describe the poetic structure of the sapiential unit.

The development of the formal structure, the linguistic structure and of the poetic structure had first to do with the description of a formal and logico-semantic model, the type of model proposed by Fernando Bello in 1981 and quoted by Salah Mejri in his book intitled Le figement lexical (1997). But it is an idealized model, which can't be used as such in computational modeling. This is why we have decided to encode differently what belongs to the form and the meaning, what is intracultural and intercultural. This explains the use of tags indicating: ideological keywords <key._>, figurative meaning <sf._>, literal meaning <sl._>, the lesson <lec._>.

We also take into account diachronic and areal aspects in appended databases concerning the works from which we extract the brief sapiential units, the authors, translators, compilers that are to be found in the encyclopedic index cards coming from critical introductions, prosopographies. All this information is directly linked to the encoded texts.

## Using the specific polysemy of brief sapiential units

In the description of the knowledge container, we mentioned the specific computational use of some XML entities, like the <lem.es> entity for example.

But we can define also the knowledge container as a computational container, where some XML entities are used to drive specific semantic calculations.

This dual conception of the knowledge representation is mentioned by *Markman* in his analysis of the requirements of a model of knowledge representation:

"It makes no sense to talk about representations in the absence of processes... Only when there    is also a process that use the representation does the system actually represent, and the    capabilities of a system are defined only when there is both a representation and a process".   [Markman]

Ultimately, the container can be thought as a dialogical resource shared as well by specialists than by any computing process [Gronoff]. This dialogic process allows to improve and validate XML knowledge representation in order to compute similarities between brief sapiential units characterized by several kinds of irregularities or between multilingual search for similarities [Wierzbicka].

Many sapiential units integrate in their linguistic representation phonetic elements to make the unit efficient to give an advice, to produce great-sounding effect when it is declaimed, to help memorisation, and /or to make puns. A great sapiential unit integrates the whole – good advice, musical and poetic declamation, replay and enjoyment – in a specific textual blending.

This kind of transcription can be analyzed as a polysemic system depending on the quality of the blending. A high quality blending is an efficient polysemic system where the borders of the different components of the system are melted and allow to slide between meaning, aesthetics and ergonomy. The quality of the polysemic system can be thought as a part of the success story of some sapiential units.

Such blended polysemic systems can be analyzed with the use of specific formal conventions working at the scale of the construal unit - in our case the brief sapiential unit -,  instead of at the scale of the words, as it is the case with statistiscal methodologies or, for instance,  in [Arbesman S.; Strogatz, S. H.; Vitevitch, M.], although the authors deal with "the presence of many smaller components (referred to as islands)..." to compare phonologically and semantically Spanish and English.

Hikam et ses sources. Journée d'étude Aliento (11 juin 2010). Inalco 201.